\documentclass[lettersize,journal]{IEEEtran}
\usepackage{amsmath,amsfonts}
\usepackage{algorithmic}
\usepackage{algorithm}
\usepackage{array}
\usepackage[caption=false,font=normalsize,labelfont=sf,textfont=sf]{subfig}
\usepackage{textcomp}
\usepackage{stfloats}
\usepackage{url}
\usepackage{verbatim}
\usepackage{graphicx}
\usepackage{cite}
\usepackage{diagbox}

\usepackage{orcidlink}

\begin{document}

\title{Common Subexpression-based Compression and Multiplication of Sparse Constant Matrices}


\author{Emre Bilgili \orcidlink{0000-0003-2341-5180 } , Arda Yurdakul \orcidlink{0000-0001-7132-0042}, \textit{Computer Engineering Department, Bogazici University}}

\maketitle
\begin{abstract}
In deep learning inference, model parameters are pruned and quantized to reduce the model size. Compression methods and common subexpression (CSE) elimination algorithms are applied on sparse constant matrices to deploy the models on low-cost embedded devices. However, the state-of-the-art CSE elimination methods do not scale well for handling large matrices. They reach hours for extracting CSEs in a $200 \times 200$ matrix while their matrix multiplication algorithms execute longer than the conventional matrix multiplication methods. Besides, there exist no compression methods for matrices utilizing CSEs. As a remedy to this problem, a random search-based algorithm is proposed in this paper to extract CSEs in the column pairs of a constant matrix. It produces an adder tree for a $1000 \times 1000$ matrix in a minute. To compress the adder tree, this paper presents a compression format by extending the Compressed Sparse Row (CSR) to include CSEs. While compression rates of more than $50\%$ can be achieved compared to the original CSR format, simulations for a single-core embedded system show that the matrix multiplication execution time can be reduced by $20\%$. 

\end{abstract}


\section{Introduction}
Tiny machine learning (TinyML) mainly targets on-device data analytics on extremely resource-constrained embedded platforms \cite{refTiny}. Most deep learning (DL) inferences perform a series of Constant Matrix Multiplication (CMM). Several methods have been proposed to speed up the process and reduce power consumption \cite{refDLHardware}. In one approach, the constant matrices are pruned at the cost of some accuracy loss. Then, the resulting sparse matrices are compressed to eliminate the processing with zero operands \cite{refMultilevelCSR}. They are recorded into several one-dimensional arrays, which are used in CMM \cite{refSparseMatrix}. Another approach utilizes quantization to reduce the number of unique entries and increase the number of duplicate elements. 
Then, common subexpression (CSE) elimination algorithms are applied to remove the redundant operations \cite{refPowerOfTwo}.

Existing CSE extraction methods are not utilized to compress sparse matrices. 
Hence, though they greatly reduce the number of operations, a resulting representation is beneficial as long as it is implemented as an accelerator \cite{refCSEDSP}. As a CSE-reduced CMM still contains zeroes, its implementation is inefficient on a resource-constrained embedded platform. Based on this observation, this paper proposes a CSE extraction method and its CSE-reduced constant matrix compression format for efficient CMM implementations. The method relies on a heuristic-based search algorithm. The search space is limited to only size-of-two CSEs to reduce the extraction time. The lossless compression method is adapted from Compressed Sparse Row (CSR) format \cite{refSparseMatrixCSR}. 
The proposed method compresses significantly more than CSR when the matrix weights are limited to a few numbers. CMM time also reduces compared to CMM with CSR. 

The paper is organized as follows. The next section summarizes the implementation styles of DL inferences. Section III describes the proposed method. The experiments are presented and discussed in Section IV. The final section concludes the work.


\section{Related Work}


Most studies targeting low-cost platforms aim to reduce the latency, power consumption and size of the inference while keeping the accuracy above a certain threshold. The solutions for reducing the costs may be grouped into four. The first group studies reshaping. The large layers are approximately compressed or decomposed into smaller ones to remove the redundant operations and reduce the size \cite{b5}. Secondly, some matrix elements are set to zero during the training step or after the weights are produced \cite{b6}. Pruning reduces each matrix density by a  predefined ratio. Thirdly, the weights are quantized to reduce the storage \cite{b7}. If the hardware contains the processing unit for the target data type, the latency is also reduced. 
The fourth group compresses the pruned matrices to ignore zero elements \cite{b8}. These methods store the non-zero elements into several one-dimensional arrays. The compression is lossless. Moreover, its two-dimensional form can be reconstructed during CMM by directly processing the one-dimensional arrays to produce the result vector. 

CSE extraction is a widely studied topic in circuit minimization. Both exact and approximate methods exist since it is an NP-complete problem \cite{refYurdakulJVLSI99}. Two efficient algorithms introduced in \cite{b9} and \cite{b10} are used to accelerate the deep learning inferences on hardware \cite{refUnrollingTNN, refMicroAI}. 
Both methods use the same notation to express CSEs: a row-column pair is appended to the matrix for each CSE. As the number of CSEs increases, the matrix expands in both dimensions, resulting in a longer CSE generation time. Hence, the run-time of these methods does not scale well for large matrices. Besides, the size of the resulting matrix requires a longer processing time and bigger storage on a single-core embedded system.
\section{METHOD}
\label{chapter:method}

Let $\mathbf{T}$ be an $M \times N$ constant matrix where a column is accessed as $\mathbf{t}_j$. Let $\mathbf{v}$ represent an input vector, and $v_j$ is one of its entries. Then, a matrix-vector multiplication can be realized as 
\begin{equation}\label{matvec}
    \mathbf{Tv} = \mathbf{y} = \sum_{j=0}^{M-1} \mathbf{t}_jv_j.
\end{equation}
Equation \ref{matvec} does matrix-vector multiplication in two steps. Firstly, each entry multiplies the related column. Then, the multiplied columns are added row-by-row. Hence, the number of multiplications is reduced when a column contains a constant more than once. When only non-zero elements contribute to CMM, the upper bound on the number of computations is determined by the number of non-zero elements, $E$. Sparsity increases as the non-zero ratio, $\alpha$, reduces:
\begin{equation}\label{NZR}
   \alpha = \frac{E}{M.N}.
\end{equation}

Computations can be reduced further if the non-zero elements are represented by only a few different numbers. For example, a matrix that contains only $-1$, $0$ and $1$ strips off the multiplications. A matrix of size $1000 \times 1000$ with $4-$bit fixed point entries requires at most 16000 multiplications instead of one million if the computation is carried out as shown in Equation (\ref{matvec}). Without loss of generality, it can be claimed that the number of unique values ($U$) in a matrix is essential in reducing the number of computations, regardless of the data format used in representing these values. 


The proposed method consists of common-subexpression extraction and compression steps. The number of multiplications and additions is reduced in the first step. Then, the constant matrix is losslessly compressed into several one-dimensional arrays in the second step. 
\subsection{Common Subexpression (CSE) Elimination}

The proposed approach searches two-element common subexpressions. Let $\mathbf{t}_i$ and $\mathbf{t}_j$ be two selected columns from matrix $\mathbf{T}$. The elements at the $r$'th row of these columns are accessed as $t_{r,i}$ and $t_{r,j}$. Element-wise addition at these rows, $add_{r,ij}$ is defined as
\begin{equation}\label{addel}
   add_{r,ij}=t_{r,i}v_i+t_{r,j}v_j.
\end{equation}
If there exists another row $q$ such that $add_{r,ij}=add_{q,ij}$, then one of the additions can be eliminated since the result of the first addition can be used in the other. Assume that there are $z_{r,ij}$ occurrences of $add_{r,ij}$ in the $(\mathbf{t}_i,\mathbf{t}_j)$ pair. Then, the number of addition eliminations due to $add_{r,ij}$ can be computed as $z_{r,ij}-1$. Thus, a solution that maximizes $gain$ is sought:
\begin{equation}
\label{gain}
gain=\sum_{\forall(\mathbf{t}_i,\mathbf{t}_j)} \sum_k (z_{r,ij}-1).
\end{equation}

\begin{algorithm}
\begin{algorithmic}[1]
\renewcommand{\algorithmicrequire}{\textbf{Input:}}
\renewcommand{\algorithmicensure}{\textbf{Output:}}
\REQUIRE {$\mathbf{T}$}
\ENSURE {$CSE, \mathbf{T}^u$}
\STATE $gain = 0$
\STATE $CSE = \emptyset $
\FOR{ every iteration $< It$}  \label{iterationstart}
	\STATE $Commons = \emptyset $ \label{initialstart}
	\STATE Form $T_a=\{\mathbf{t}_0,\mathbf{t}_1, \dots, \mathbf{t}_n\}$
	\STATE partition $T_a$ in to $N/2$ pairs: $\mathbf{t_{ij}}=(\mathbf{t}_i,\mathbf{t}_j)$
	\FOR {every $t_{ij}$ across all $r$ rows} 
		\STATE find $add_{r,ij}$ such that $z_{r,ij}>1$ is maximum for $t_{ij}$
     	\ENDFOR
	\STATE $gain = \sum_{\forall t_{ij}}\sum_r (z_{r,ij}-1)$
     	\STATE $Commons = \{add_{r,ij}$ that satisfy current $gain\}$ \label{initialend}
	\FOR {every attempt $< At$} 
     		\STATE dissolve any $t_{ij}$ and $t_{kl}$ to obtain $t_{ik}$ and $t_{jl}$ \label{attemptstart}
		\FOR {the new pairs across all $r$ rows} 
			\STATE find $add_{r,ij}$ such that $z_{r,ij}>1$ is maximized
     		\ENDFOR
		\IF {$gain \leq \sum_{\forall t_{ij}}\sum_r (z_{r,ij}-1))$}
     			\STATE update $gain$ and $Commons$ with  $t_{ik}$ and $t_{jl}$
		\ELSE \STATE {restore pairs as $t_{ij}$ and $t_{kl}$}
		\ENDIF \label{attemptend}
  	\ENDFOR 
	\STATE Update T by removing copies of $add_{r,ij} \in Commons$
  	\STATE Update $CSE$ by including $Commons$
\ENDFOR \label{iterationend}
\STATE $T^u=T$ 
\end{algorithmic}
\caption{Algorithm for CSE detection}
\label{AlgCSE}
\end{algorithm}

\begin{figure}
\centering{\includegraphics[width=0.7\columnwidth]{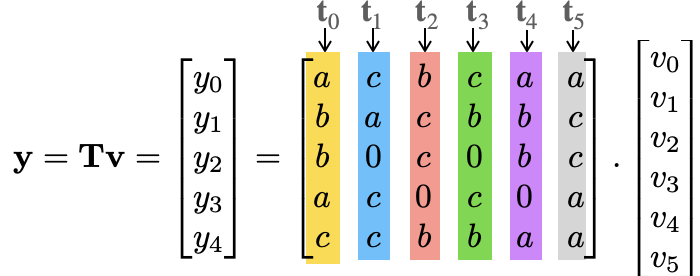}}
\centerline{(a)}
\raggedright
\begin{footnotesize}
\textbf{Start of an iteration}
\begin{itemize}
    \item $(\mathbf{t}_0,\mathbf{t}_3)$
    : $av_0 + cv_3$ occurs twice.
    \item $(\mathbf{t}_1,\mathbf{t}_4)$
    : $cv_1 + av_4$ occurs twice.
    \item $(\mathbf{t}_2,\mathbf{t}_5)$
    : $bv_2 + av_5$ and $cv_2 + cv_5$ occur twice.
\end{itemize}
$gain=4$. $Commons=\{(\mathbf{t}_0,\mathbf{t}_3), (\mathbf{t}_1,\mathbf{t}_4),(\mathbf{t}_2,\mathbf{t}_5)\}$.\\
\textbf{First attempt: }Pick $(\mathbf{t}_0,\mathbf{t}_3)$ and $(\mathbf{t}_2,\mathbf{t}_5)$. Exchange $\mathbf{t}_2$ and $\mathbf{t}_3$.
\begin{itemize}
    \item $(\mathbf{t}_0,\mathbf{t}_2)$
    : $bv_0 + cv_2$ occurs twice.
    \item $(\mathbf{t}_1,\mathbf{t}_4)$
    : $cv_1 + av_4$ occurs twice.
    \item $(\mathbf{t}_3,\mathbf{t}_5)$
    : $cv_3 + av_5$ occurs twice.
\end{itemize}
$gain=3$. Restore $(\mathbf{t}_0,\mathbf{t}_3)$ and $(\mathbf{t}_2,\mathbf{t}_5)$.\\
\textbf{Second attempt: }Pick $(\mathbf{t}_1,\mathbf{t}_4)$ and $(\mathbf{t}_2,\mathbf{t}_5)$. Exchange $\mathbf{t}_4$ and $\mathbf{t}_5$.
\begin{itemize}
    \item $(\mathbf{t}_0,\mathbf{t}_3)$
    : $av_0 + cv_3$ occurs twice.
    \item $(\mathbf{t}_1,\mathbf{t}_5)$
    : $cv_1 + av_5$ occurs three times.
    \item $(\mathbf{t}_2,\mathbf{t}_4)$
    : $bv_2 + av_4$ and $cv_2 + bv_4$ occur twice.
\end{itemize} 
$gain=5$. $Commons=\{(\mathbf{t}_0,\mathbf{t}_3), (\mathbf{t}_1,\mathbf{t}_5),(\mathbf{t}_2,\mathbf{t}_4)\}$.\\
\end{footnotesize}
\centerline{(b)}
\caption{Two-term CSE extraction: (a)Example matrix, (b)Sample iteration of Algorithm \ref{AlgCSE}.}
\label{FigExampleAlg}
\end{figure}

The search algorithm consists of a series of iterations (Algorithm \ref{AlgCSE}). Each iteration includes \textit{initial} and \textit{improvement} phases. In the initial phase, $N$ columns are randomly paired to form $\frac{N}{2}$ column pairs (lines \ref{initialstart}-\ref{initialend}). The improvement phase consists of a series of \textit{attempts}. In each attempt, two random columns from different pairs are temporarily exchanged for possible gain improvements (lines \ref{attemptstart}-\ref{attemptend}). At the end of an iteration, the common subexpressions are eliminated from the matrix. The next iteration runs on this updated matrix. The number of iterations ($It$) and the number of attempts ($At$) per iteration can be decided by the user though it can also end when gain does not decrease in successive iterations or no more addition pairs with $z_{r,ij}>1$ can be found. The algorithm returns a set of common subexpressions and the final update of the matrix, $T^u$. A sample iteration is shown in Fig. \ref{FigExampleAlg}. 

\subsection{CSE-based Matrix Compression}
The CSE detection algorithm returns the set of two-term CSEs and the remainder matrix $\mathbf{T}^u$. We propose three pairs of one-dimensional arrays to store the original matrix $\mathbf{T}$. In each pair, the first array stores the values, and the second array stores the pointers to process the first array. The last element of the second array holds the length of the first array.
\subsubsection {Weights}
The first array stores nonzero entries in each column of the matrix as values in $U$ because each entry of the input vector is multiplied with the weights of the column only once. The second array $WP$ shows the start index of the weight set for the next column. In Fig. \ref{FigExampleAlg}(a), $a$ and $b$  appear twice in $\mathbf{t}_0$. Fig. \ref{FigExampleArr}(a) represents them as one $a$ and one $b$ in the yellow slots. They are multiplied by $v_0$ only once. $WP$ starts with 3, indicating the start of the unique weights for $\mathbf{t}_1$.

\subsubsection {CSE}
The elements of the CSE set obtained in Algorithm \ref{AlgCSE} are stored with their row positions in the first array. The second array $CP$ shows the start index of the next CSE. In Fig. \ref{FigExampleAlg}, $av_0+cv_3$ is selected as a CSE at the end of the iteration. It is composed of $a$ in the yellow slot and $c$ in the green slot of \textit{Weights} in Fig. \ref{FigExampleArr}(a). Their indices are 0 and 8 in the \textit{Weights} array, respectively. This information is stored as the first two entries in the \textit{CSE} array in Fig. \ref{FigExampleArr}(b). The next two slots are reserved for the rows where this CSE appears, because \textit{CP} begins with 4 as the start index of the next CSE. Fig. \ref{FigExampleAlg}(a) shows that $av_0+cv_3$ is at rows 0 and 3. Hence $CSE[2]=0$, $CSE[3]=3$ in Fig. \ref{FigExampleArr}(b). 

\subsubsection {Singles}
\begin{figure}
\centering
\includegraphics[width=0.7\columnwidth]{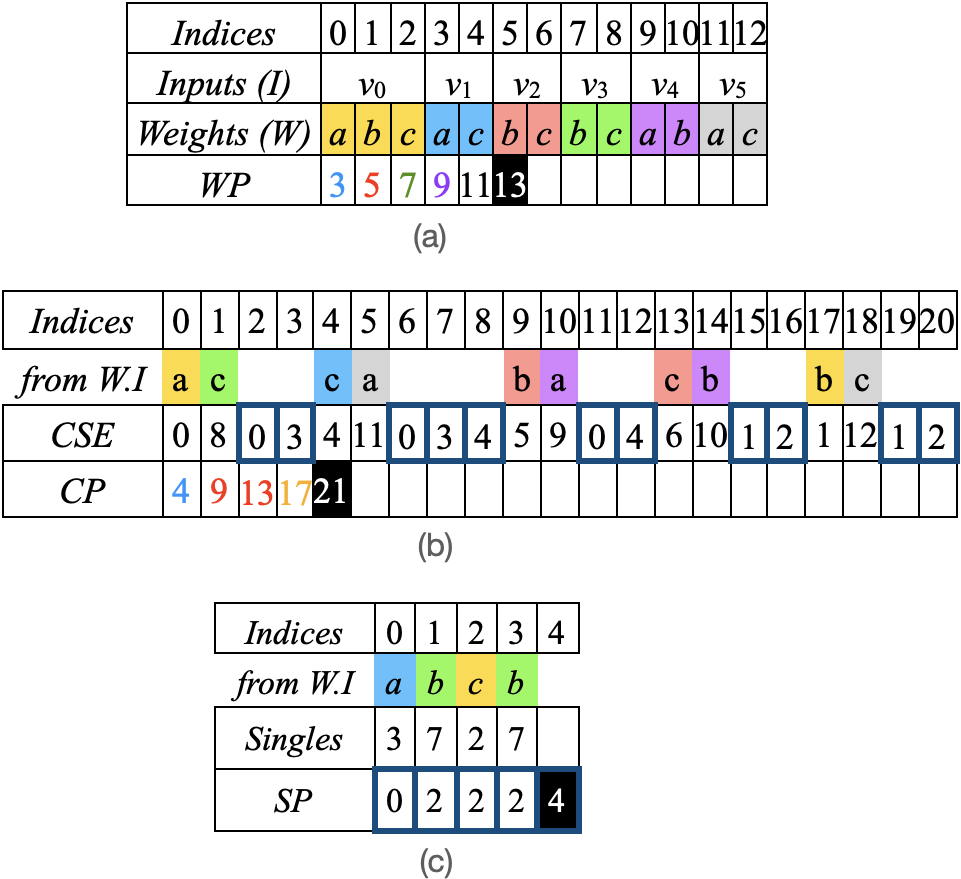}
\caption{Array pairs for matrix compression: (a) weights, (b) CSE, (c) Singles}
\label{FigExampleArr}
\end{figure}

The elements of the remainder matrix $\mathbf{T}^u$ are stored one by one in the first array. Fig. \ref{FigExampleArr}(c) shows $av_1$ as the first entry. It is the third entry from the \textit{Weights} array. Its row 
in $\mathbf{T}^u$ is given in $SP$ array. $SP[0]$ shows the number of elements in row 0. $SP[r]-SP[r-1]$ shows the number of elements in row $r$. As $SP[0]=0$ and $SP[1]=2$ in Fig. \ref{FigExampleArr}(c), row 0 is empty and row[1] has two elements, $av_1$ and $bv_3$.

\subsection{Analysis}
\begin{figure}
\centering
\includegraphics[width=1.0\columnwidth]{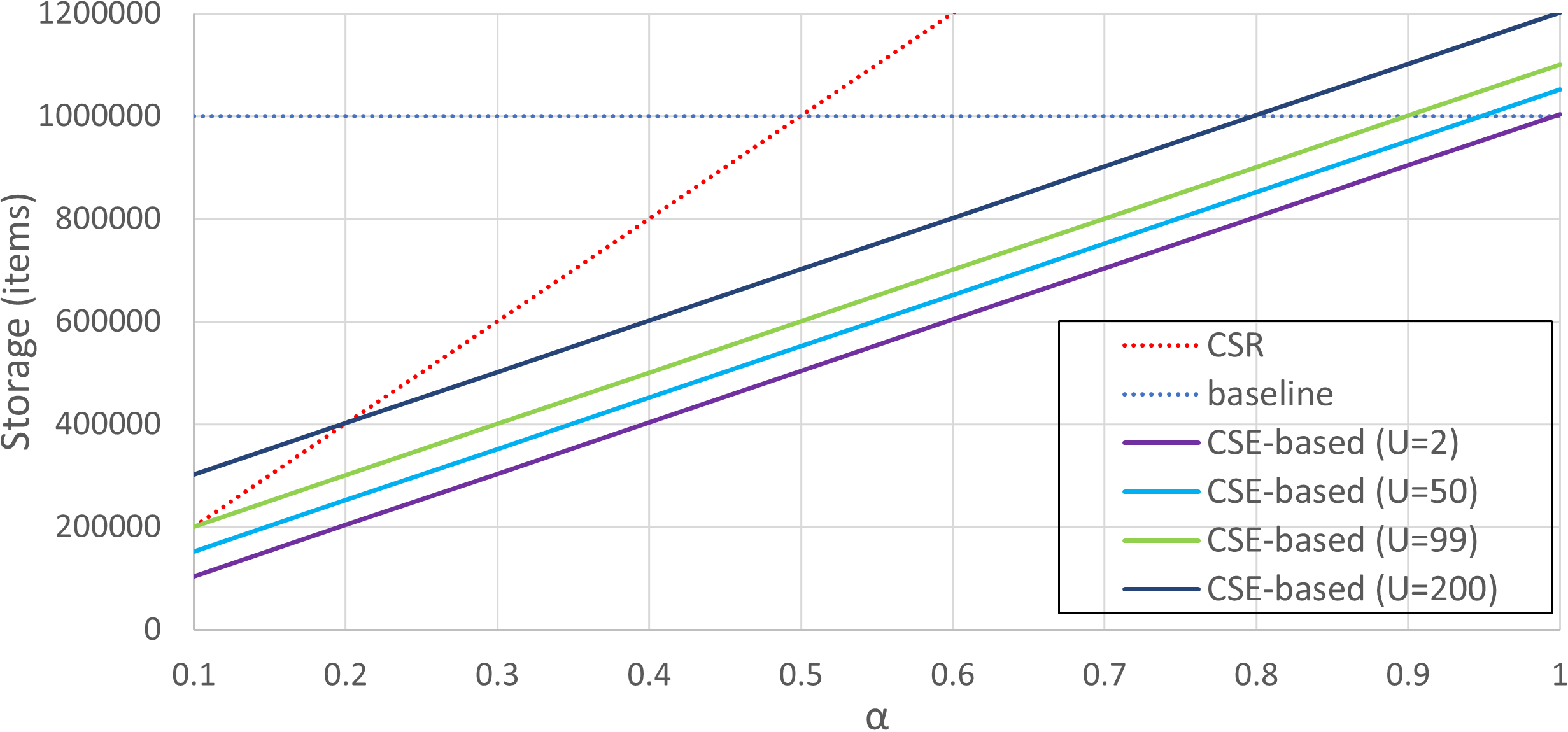}
\caption{Storage Comparison of CSR, Baseline and the Proposed Method.}
\label{FigCSRvsMNvsProp}
\end{figure}

In Compressed Sparse Row (CSR) representation, only nonzero elements with their column positions are stored. The row information is obtained in the same way that we use in the SP of the Singles array
. Then, the storage cost of CSR is 
\begin{equation}\label{eqCSR}
S_{CSR}=2. E+M.
\end{equation}
Using Equation \ref{NZR}, it can be deduced that CSR stores a matrix in a smaller memory when the following condition holds:
\begin{equation}
\alpha < \frac{N-1}{2N}.
\end{equation}
This shows CSR efficiently compresses large matrices if sparsity is more than 50\%. Our proposal offers more compression when the number of unique values is limited. The storage consumed by  \textit{Weights} and \textit{WP} is 
\begin{equation}
S_{Weights}\leq N.(U+1).
\end{equation}
The worst case occurs when all unique values appear in each column. The \textit{CSE} array pair requires two entries for each common subexpression in the CSE set of Algorithm \ref{AlgCSE}. The number of outputs that use the common subexpressions is $gain+ |CSE|$. The \textit{CP} array size is $|CSE|$. Then,
\begin{equation}
S_{CSE}=gain+ 4. |CSE|
\end{equation}
In the \textit{Singles} array pair, the first array has $E-2.(gain+|CSE|)$ elements. Including the number of rows, the storage cost is 
\begin{equation}
S_{Singles}=E-2.(gain+|CSE|)+ M.
\end{equation}
Then, the total cost of the proposed approach is
\begin{equation}\label{eqProposed}
S_{prop} \leq N.(U+1)+E+M+2.|CSE|-gain
\end{equation}
Equations \ref{eqCSR} and \ref{eqProposed} can be processed to show that the proposed approach compresses better than CSR when the following condition is satisfied:
\begin{equation}\label{eqCriterion}
\alpha > (U+1)/M+(2.|CSE|-gain)/(M.N)
\end{equation}
Using Equation \ref{eqCriterion}, Fig. \ref{FigCSRvsMNvsProp} is plotted to show the storage for a $1000\times1000$ matrix  with $|CSE|=0$ for different $\alpha$ and $U$ values. When $\alpha=0.1$ and $U=99$, the proposed approach compresses exactly the same as CSR. However, its compression is better than CSR as $U$ gets lower. Especially for binary or ternary matrices, compression is still achieved with the new scheme even though the matrix is not sparse. 
Equation \ref{eqCriterion} also shows that when the average $gain$ per common subexpression is more than two, the number of unique values can be increased further.

\section{Experiments}

Sample matrices are constructed to compare the baseline, CSR, \cite{b9}, \cite{b10} and the proposed method. Sparse matrices are obtained by removing the lowest absolute valued-entries until the desired $\alpha$ is attained \cite{refPruningQuantization}. Then, nonzero entries are linearly quantized with $U$ distinct values. The baseline consists of a nested for loop to multiply and accumulate nonzero entries. The matrix multiplication with each method is simulated on gem5 \cite{b13} for a single x86 ISA CPU running at 1 GHz, two levels of 256MB cache implemented as a single-port 

Methods in \cite{b9} and \cite{b10} can process only 0-1 matrices. Hence, only the number of additions for matrices with $M=N=100$, $U = 2$ and $\alpha = \{0.25, 0.5, 0.75\}$  are listed in Table \ref{TabResults}. 
The matrix size is limited to 100 as \cite{b9} and \cite{b10} produce an adder tree in an hour. The proposed method consumes 70 seconds for the same matrix. As $\alpha$ decreases, the length of the CSEs decreases, and the proposed method produces better results. When the length of the CSEs is allowed to increase, \cite{b9} and \cite{b10} can yield fewer additions as their CSEs can include multiple terms. However, the CPU cycles for CMM using \cite{b9} and \cite{b10} are $2.0\times$--$4.7\times$ and $1.6\times$--$2.0\times$ of the proposed method as they lack compression. Execution time of CMM with CSR slightly exceeds the proposed method.

\vspace*{-.2cm}

\begin{table}[htbp]
\caption{The Number of Additions for $U = 2$ and $M=N=100$}
\begin{center}
\begin{tabular}{|c||c|c|c|c|c|}
\hline
$\alpha$ & CSR & baseline & \cite{b9} & \cite{b10} & CSE-based \\
\hline \hline
0.25 & 2500 & 2500 & 2093 & 2072 & \textbf{1923} \\
\hline
0.5 & 5000 & 5000 & 3353 & 3537 & \textbf{3326} \\
\hline
0.75 & 7500 & 7500 & \textbf{4229} & 4536 & 4492 \\
\hline
\end{tabular}
\label{TabResults}
\end{center}
\end{table}

\vspace*{-.2cm}

\begin{figure}[htbp]
\centering
\includegraphics[width=1.0\columnwidth]{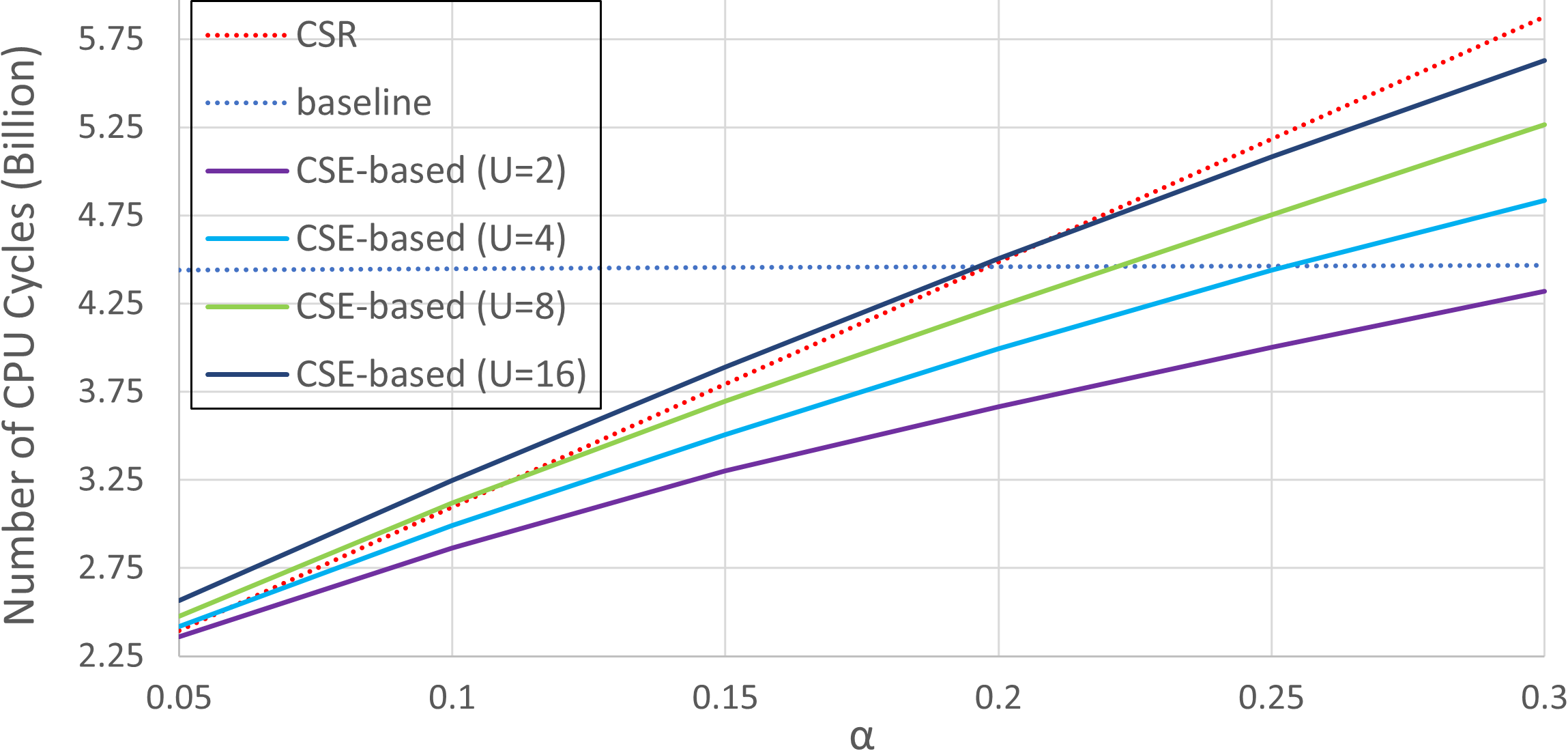}
\caption{CMM Computation time comparison of CSR, Baseline and the Proposed Method.}
\label{FigLatency}
\end{figure}

The performance of the proposed method, baseline and CSR are compared for $M=N=1000$ and $\alpha \leq  0.3$. Fig. \ref{FigLatency} shows that CSR  performs better with lower $\alpha$ while the proposed method is better with lower $U$ and higher $\alpha$.

\vspace*{-.2cm}

\vspace*{-.2cm}

As a DL use case, an autoencoder is trained with the MNIST dataset for 98.85\% accuracy. It contains eleven matrices from $64 \times 128$ to $2048 \times 4096$. Nine versions of this autoencoder are prepared with $U=\{2,4,8\}$ and $\alpha=\{0.25,0.50,0.75\}$. 
In Table \ref{TabAutoencoder}, the ratio of the total storage size over the baseline $(N \times M)$ is presented. The results show that a matrix represented with a small unique value set is compressed better with the proposed method than it is done CSR. The storage size decreases when $U$ decreases in the proposed method. 

\begin{table}[htbp]
\caption{Autoencoder Storage Size Relative to the Baseline}
\begin{center}
\begin{tabular}{|c||c|c|c|c|}
\hline
$\alpha$ & CSR & CSE ($U = 2$) & CSE ($U = 4$) & CSE ($U = 8$) \\
\hline \hline
0.25 & 50.1\% & \textbf{22.5\%} & 23.4\% & 24.6\% \\
\hline
0.5 & 100.1\% & \textbf{38.2\%} & 39.4\% & 42.5\% \\
\hline
0.75 & 150.1\% & \textbf{47.5\%} & 48.8\% & 52.8\% \\
\hline
\end{tabular}
\label{TabAutoencoder}
\end{center}
\end{table}

\vspace*{-.2cm}

\section{Conclusion}

A fast CSE extraction algorithm with a dedicated compression format is proposed for large sparse matrix multiplication. It compresses more than CSR when weights are represented with a few numbers. Simulations show that it outperforms the state-of-the-art methods for this type of large matrices. 

\clearpage

\end{document}